%% file: main_pscc_final.tex
\documentclass{IEEEtran4PSCC}

\makeatletter
\let\old@ps@headings\ps@headings
\let\old@ps@IEEEtitlepagestyle\ps@IEEEtitlepagestyle
\def\psccfooter#1{%
    \def\ps@headings{%
        \old@ps@headings%
        \def\@oddfoot{\strut\hfill#1\hfill\strut}%
        \def\@evenfoot{\strut\hfill#1\hfill\strut}%
    }%
    \def\ps@IEEEtitlepagestyle{%
        \old@ps@IEEEtitlepagestyle%
        \def\@oddfoot{\strut\hfill#1\hfill\strut}%
        \def\@evenfoot{\strut\hfill#1\hfill\strut}%
    }%
    \ps@headings%
}
\makeatother

\psccfooter{%
        \parbox{\textwidth}{\hrulefill \\ \small{22nd Power Systems Computation Conference} \hfill \begin{minipage}{0.2\textwidth}\centering \vspace*{4pt} \includegraphics[scale=0.06]{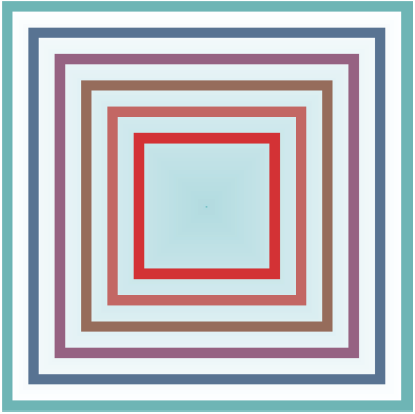}\\\small{PSCC 2022} \end{minipage} \hfill \small{Porto, Portugal --- June 27 -- July 1, 2022}}%
}

\usepackage{xspace,empheq,fancybox,amsmath,amssymb,graphicx,epstopdf,epsfig,subfigure,syntonly,times,amsthm} \usepackage{psfrag,color,xcolor,bm,array,cite,bbm}
\usepackage{url,cite,footnote,xspace,syntonly,algorithm,algpseudocode}
\usepackage{verbatim,multirow}
\usepackage[T1]{fontenc}
\usepackage{multirow,flushend}

\newtheorem{remark}{\bfseries Remark}
\input{mysymbol.sty}

\newcommand{\T}{\mathsf{T}}

\begin{document}
%
\title{Risk-Aware Learning for Scalable Voltage Optimization in Distribution Grids}

\author{
\IEEEauthorblockN{Shanny Lin, Shaohui Liu, and Hao Zhu}
\IEEEauthorblockA{Department of Electrical and Computer Engineering  \\
The University of Texas at Austin\\
\{shannylin, shaohui.liu, haozhu\}@utexas.edu}
}

\maketitle

\begin{abstract}
Real-time coordination of distributed energy resources (DERs) is crucial for regulating the voltage profile in distribution grids. By capitalizing on a scalable neural network (NN) architecture, one can attain decentralized DER decisions to address the lack of real-time communications. This paper develops an advanced learning-enabled DER coordination scheme by accounting for the potential risks associated with reactive power prediction and voltage deviation. Such risks are quantified by the conditional value-at-risk (CVaR) using the worst-case samples only, and {we propose a mini-batch selection algorithm to address the training speed issue in minimizing the CVaR-regularized loss.}
Numerical tests using real-world data on the IEEE 123-bus test case have demonstrated the computation and safety improvements of the proposed risk-aware learning algorithm for decentralized DER decision making, {especially in terms of reducing feeder voltage violations.}
\end{abstract}


\thanksto{\noindent This work has been supported by NSF Grants 1802319 and 2130706. The first two authors have contributed equally to this work.} 


\input{Introduction}

\input{Modeling}

\input{LearningModel}
\input{Results}

\input{Conclusion}

%
\bibliographystyle{IEEEtran}
\itemsep2pt
\bibliography{IEEEabrv,hzpub,References}

\end{document}

%% file: Introduction.tex
\section{Introduction} \label{sec:Intro}

Rapid integration of distributed energy resources (DERs) opens up new opportunities of flexible and adaptive support to the operations of power distribution grids. The smart inverters of fast-acting DERs have been popularly advocated for the feeder voltage optimization by adjusting their reactive power outputs \cite{IEEE1547}. It is crucial to design a scalable coordination framework for these heterogeneous DERs under limited real-time communications available in distribution systems.

Coordinating reactive power setpoints of DERs can be viewed as an optimal power flow (OPF) problem that requires all feeder-wide information. To bypass the high communication overhead of a centralized solution, several distributed and decentralized optimization algorithms have attracted significant interest in the last years; see e.g., {\cite{molzahn2017survey,antoniadou2017distributed, liu2017distributed}}.  
{ While those algorithms can greatly simplify the communication graph to have information exchange only among neighbors or with the control center, they still require high-rate bi-directional communications to cope up with the fast dynamics in distribution grids.}
{With {the} increasing volume of available data, the recent trend is to leverage machine learning (ML) tools to attain decentralized decision rules that map from {the} system operating condition (OC)  to the optimal decisions {\cite{jalali2020designing,gupta2021controlling,yang2020two,dobbe2020toward,cao2021deep}.} By and large, these ML-based approaches have been designed to reduce the sample average of prediction error, or the \textit{average loss}. Accordingly, they may fail to address the {largest sample losses or} worst-case scenarios in terms of prediction error or even the violation of voltage limits.   

Our goal is to develop a risk-aware learning framework to improve the safety of scalable decision rules developed for distribution grid voltage optimization. Inspired by earlier work {\cite{jalali2020designing,gupta2021controlling,yang2020two,dobbe2020toward,cao2021deep}}, we design a scalable neural network architecture such that each nodal inverter can learn its optimal reactive power from local measurements and selected information only. To address the aforementioned issue of optimizing average loss only, {it is possible to adopt a post-analysis approach that reduces the worst-case errors of the resultant NN models by 
restricting the input domain \cite{Venzke2020worstcase}.  
Alternatively,} we 
{propose to \textit{systematically} incorporating the worst-case measure into the training process by introducing a regularization term on} the \textit{conditional value-at-risk (CVaR)} \cite{rockafellar2000optimization} of the predicted decisions. 
The CVaR metric corresponds to the average of the largest sample losses, or the worst-case scenarios, and thus optimizing it can improve the safety associated with the reactive power decisions. Specifically for our voltage optimization task, we also use the CVaR metric to quantify the worst-case voltage deviation performance. {This way, the NN design is also physics-informed by incorporating the system-wide voltage limits; see .e.g, \cite{nellikkath2021physics}.}
Notably, the CVaR metric is defined over a subset of samples (worst ones) and thus can suffer from the learning efficiency issue. We further propose to accelerate the mini-batch gradient descent for NN training \cite{nemirovski2009robust,ghadimi2016mini} by selecting those mini-batches attaining a certain CVaR threshold. {Accordingly, gradient updates are performed only on the mini-matches of statistical significance, thus reducing the number of epochs.} Our algorithm has effectively improved the computation time for the training process and reduced the voltage violations based on numerical tests.

It is worth mentioning the proposed risk-aware learning framework is very related to a recent work \cite{gupta2021controlling} on addressing the CVaR risk of voltage deviations in OPF-based NN training. While this paper incorporates CVaR in a stochastic optimization framework, our proposed learning approach here directly uses the scenarios generated by OPF solutions. Thus, the present work can be easily generalized to incorporate other operational considerations in real-time feeder decision making. Notably, we also present an algorithmic solution to effectively accelerate the NN training process. }


The rest of this paper is organized as follows. Section \ref{sec:model_opt} presents the system modeling and the centralized voltage optimization problem. In Section \ref{sec:learning}, we first formulate the risk-aware learning problem by introducing the CVaR losses for prediction and voltage deviations. A mini-batch selection scheme is developed to accelerate the training process. Numerical tests using real-world data on the single-phase equivalent of the IEEE-123 bus test case are presented in Section \ref{sec:results} to demonstrate the computation and safety improvements of the proposed algorithm. The paper is wrapped up in Section \ref{sec:conclusion}. 

\textit{Notation:} Upper (lower) boldface symbols stand for matrices (vectors); $(\cdot)^{\T}$ stands for matrix transposition; $\|\cdot\|_2$ denotes the $L_2$-norm; $|\cdot|$ denotes the absolute value; $\nabla_x$ denotes the gradient with respect to $x$; $\mathbbm{1}$ denotes the indicator function; and $\mathbf 0$ stands for an all-zero vector of appropriate size.

%% file: Modeling.tex
\section{System Modeling} \label{sec:model_opt} 

Consider a radial distribution feeder consisting of $(N+1)$ buses with bus $0$ denoting the reference bus at the feeder head. For simplicity, this work focuses on single-phase feeders, while results can be extended to multi-phase systems following earlier approaches such as \cite{hzml_tps16}. Let vectors $\bbp \in \mathbb R^N$ and $\bbq \in \mathbb R^N$ collect the net active and reactive power injections, respectively, at all non-reference buses. The net reactive power consists of DER generation $\bbq^g$ (controllable) and load consumption $\bbq^c$ (non-flexible) such that $\bbq := \bbq^g - \bbq^c$; similarly, the net active power injection is $\bbp = \bbp^g - \bbp^c$. 

The distribution voltage optimization task aims to coordinate the controllable $\bbq^g$ from DERs to support system operations in terms of feeder voltage regulation \cite{molzahn2017survey,antoniadou2017distributed} 
or phase balance in multi-phase systems \cite{yao2020mitigating,girigoudar2020impact}. We formulate a general centralized problem of minimizing a system-wide operational objective in order to satisfy 
voltage limit constraints, 
while adhering to reactive power limits $\mathcal Q$, as 
\begin{subequations} \label{eqn:LCQP}
\begin{align} 
{\bbz} = & \min_{\bbq^g \in \mathcal Q}&  &o(\bbq^g) \label{eqn:LCQP_cost}\\ 
& \text{s. to}&  &\bbg(\bbq^g) \leq \mathbf 0 \label{eqn:LCQP_constraint}
\end{align} 
\end{subequations} 
where 
${\bbz}$ denotes the optimal value of $\bbq^g$ by solving \eqref{eqn:LCQP}. 
To model the system-wide power flow in \eqref{eqn:LCQP}, one can simplify the accurate nonlinear power flow model by adopting the linearized DistFlow (LDF) approximation \cite{baran1989network}. The LDF model represents feeder voltage to be linear with respect to (wrt) power injections $\bbp$ and $\bbq$, and has been shown very effective for developing algorithms of distribution monitoring 
{\cite{donti2019matrix,song2019dynamic}} and voltage optimization 
{\cite{xu2019data,tang2018fast}}. The per-unit (pu) voltage \textit{deviation} from the
reference bus voltage is approximated by {the LDF model as}
\begin{align} \label{eqn:volt_dev}
\bbv \approxeq \bbR \bbp + \bbX \bbq
\end{align}
where matrices $\bbR \in \mathbb R^{N\times N}$ and $\bbX \in \mathbb R^{N\times N}$ depend on the feeder topology and line parameters. 

For the system-wide objective, one can consider the feeder ohmic loss which is quadratic wrt $\bbp$ and $\bbq$ {\cite{taheri2020fast}}. This is because the line power flow under LDF equivalently aggregates the total down-stream power injections. Thus, the objective in \eqref{eqn:LCQP_cost} becomes a convex quadratic function of $\bbq^g$, given by
\begin{align}
o(\bbq^g) = (\bbq^g)^\T \bbR \bbq^g - 2(\bbq^c)^\T \bbR \bbq^g.
\end{align}
In regard to the constraint set $\ccalQ$, each node $n$ has a reactive power limit depending on the apparent power rating and active power output of its own inverter. Collectively, the reactive power limit is set to be
\begin{align}
\mathcal Q \coloneqq \{\bbq^g: |q^g_n| \leq \bar{q}^g_n,~ \forall n = 1,\cdots, N\}.
\end{align}
Note that non-controllable nodes can be easily included too by setting the corresponding limits $\bar{q}^g_n=0$. The inequality constraint in \eqref{eqn:LCQP_constraint}
 limits all non-reference bus voltage deviations $\bbv$ to be within a fixed range $[\underline{\bbv},~\overline{\bbv}]$, by  
\begin{align} \label{eqn:safe_constraint}
\bbg(\bbq^g) = \begin{bmatrix}
\bbX\bbq^g + \bbh(\bby) - \overline{\bbv} \\
-\bbX\bbq^g - \bbh(\bby) + \underline{\bbv} \end{bmatrix} \leq \mathbf 0
\end{align} 
where $\bbh(\bby) = \bbR(\bbp^g-\bbq^c) - \bbX\bbq^c$ with $\bby:= [\bbp^c;~\bbp^q;~\bbq^c]$ capturing all system-wide inputs to problem \eqref{eqn:LCQP} for determining the feeder's operating condition (OC). 



Clearly, the voltage optimization problem \eqref{eqn:LCQP} is a linearly-constrained quadratic program (LCQP) which can be efficiently solved given the full feeder model and system-wide OC. We will introduce a risk-aware learning framework to attain scalable and safe decision rules for real-time DER operations under minimal system-wide information. Before that, the following remark discusses the generalizability of the presented models and formulation for voltage optimization.
{
\begin{remark} \label{rmk:model}
	\emph{(Multi-phase and nonlinear models.)} The optimization problem \eqref{eqn:LCQP} can be extended to multi-phase systems by using the general multi-phase LDF model; see e.g., \cite{hzml_tps16}. Basically, the network matrices $\bbR$ and $\bbX$ are formed to capture phase-to-phase connections in multi-phase lines, in a similar fashion to the single-phase case. In addition, the problem \eqref{eqn:LCQP} can be formulated using different types of linearized approximation as well, such as the fixed-point linearization {in \cite{bernstein2018load}}. 
Last, one can formulate the problem using nonlinear ac power flow model too, thanks to the popular convex relaxation based approaches {\cite{edhzgg_tsg13,low2014convex}}. 
Regardless of the underlying problem modeling, the proposed learning-enabled framework can work by using any data samples of $\{\bby, \bbz\}$ generated by solving the specific voltage optimization problem.
\end{remark} 
}

%% file: LearningModel.tex
\section{Risk-Aware Learning} \label{sec:learning}

Machine learning (ML) techniques have been recently utilized to attain scalable solutions such that individual nodes can form their own optimal decisions using minimal real-time information. The centralized problem \eqref{eqn:LCQP} is efficiently solvable but requires a wide deployment of communication resources to connect the DERs. Lacking real-time communications, each node $n$ may resort to a decentralized architecture by forming its optimal $z_n$ from local measurements and possibly limited information elsewhere. Recent work \cite{jalali2020designing,gupta2021controlling} has proposed to obtain decentralized decision rules through supervised learning approaches such as kernel learning and neural networks (NNs). {With a pre-trained model, these approaches have extremely high computation efficiency during real-time implementation.} Nonetheless, most ML-enabled approaches for end-to-end distribution system learning aim to minimize the average losses in predicting the optimal $\bbz$. This may fall short in reducing the statistical risks of the resultant decision rules. We consider a risk-aware learning framework to address this issue of ML-based DER operations. 

Specifically, the key of the ML-based solutions is to obtain the predictive model $\Phi(\bby) \rightarrow {\bbz}$ from the OC $\bby$ to the optimal $\bbz$, such that it follows a pre-specified \textit{scalable} structure. Considering fully local decision rules for example, we can enforce the model $\{\Phi_{n}(\cdot)\}_{n}$, one for each node $n$, to use local measurements only as the input features; i.e.,  
the nodal prediction is $\hat{z}_{n} = \Phi_{n} ( \bby_{n}) = \Phi_n ([p^g_{n}, p^c_n, q^c_{n}])$. Local input $\bby_n$ can also include other measurements such as nodal voltage or current magnitude. If certain real-time communications such as broadcasted messages from the feeder head or other key nodes are possible, then $\bby_n$ can also include the aggregated power flow or current magnitude in these feeder locations.  This ML-enabled framework is very scalable to large networks and flexible to varying communication scenarios.

To construct the individual mapping $\Phi_n(\cdot)$, one can adopt the NN model known for its superior nonlinear approximation capability {\cite{lecun2015deep}}. As a multi-layer perceptron (MLP), the NN model is basically a layered network with a linear transformation followed by nonlinear activation per layer. Per node $n$, using the input to the first layer $\bby_n^0$ as a vector embedding of $\bby_n$, each layer 
$t$ is given by
\begin{align}
	\mathbf{y}_n^{t+1}=\sigma(\mathbf{W}_n^t \mathbf{y}_n^t + \mathbf{b}_n^t),~~\forall t = 0,\ldots, T-1
\end{align}
where 
$\bbW_n^t$ and 
$\bbb_n^t$ are parameters to be learned, while $\sigma(\cdot)$ is the nonlinear activation such as ReLU. 
{Thus, the training processes is parallelizable as each mapping $\Phi_n(\cdot)$  can be obtained individually.} For simplicity, one can set  the parameters to be the same 
for a subset of nodes. For example, all the DER nodes within a feeder, or those DER nodes belonging to the same lateral, can use the same set of parameters.  This way, the nodes with the same parameters will be jointly trained by a single MLP. For notational simplicity, the rest of paper presents the individual NN training with  $\bbvarphi$ collecting all the learnable parameters 
$\{\bbW_n^t,\bbb_n^t\}$. Hence, there is a one-to-one mapping between the model $\Phi(\cdot)$ and parameter $\bbvarphi$, and the goal becomes to obtain the NN parameter $\bbvarphi$ . 
	
\begin{figure}[tb!]
\vspace*{-5mm}
\centering
\includegraphics[width=0.8\linewidth]{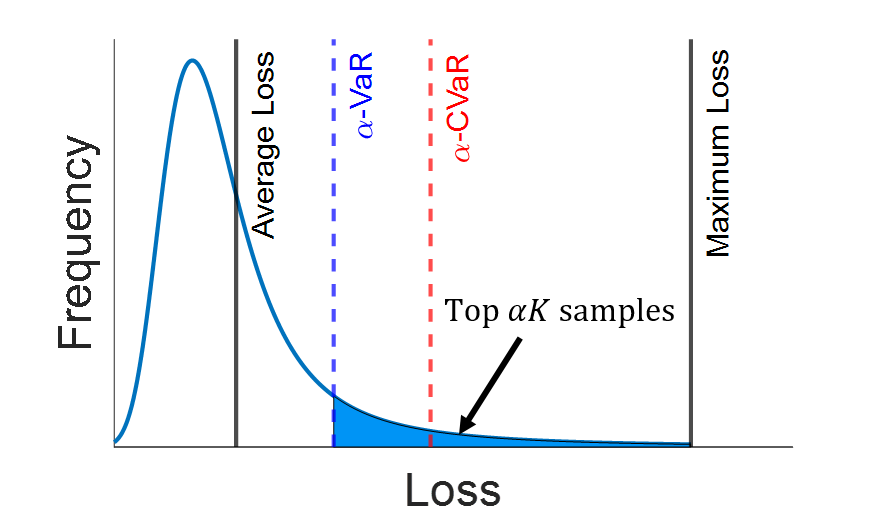}
\caption{Illustration of the $\alpha$-VaR and $\alpha$-CVaR as compared to average loss.}
\label{fig:cvar}
\end{figure}

Given the scalable structure of $\Phi(\cdot):=\{\Phi_{n}(\cdot)\}$, a centralized entity can learn $\bbvarphi$ through offline training using $K$ data samples. Each sample $k$ consists of all available measurements in $\bby_k$ as the input, and the corresponding optimal 
${\bbz}_k$ as the output. For simplicity, the rest of paper will use $k$ to index samples.
 The sample inputs can be from historic measurements,
 or the latest load forecasting, while the outputs are obtained by solving \eqref{eqn:LCQP} for each
input scenario. To learn $\bbvarphi$, a popular metric is to minimize the average loss in predicting $\bbz$ over all $K$ samples, 
as given by 
\begin{align} \label{eqn:MSE}
	 \min_{\bbvarphi} f(\bbvarphi)  :=  \textstyle \frac{1}{K}  \sum_{k = 1}^K \ell \left( \Phi(\bby_k; \bbvarphi), \bbz_k \right)
\end{align}
where $\ell (\cdot)$ denotes the loss function for each sample's predicting value based on the NN parameter $\bbvarphi$. Under the $L_2$-norm based quadratic loss given by
\begin{align*}
\ell \left( \Phi(\bby_k; \bbvarphi), \bbz_k \right) = \left\|\Phi(\bby_k; \bbvarphi) - { {\bbz}_k} \right\|_2^2,
\end{align*}
we form the mean-squared error (MSE) of prediction in \eqref{eqn:MSE}. Other error norms such as the $L_1$-norm or Huber loss can be used as well.
Due to the nonlinearity of $\Phi$ wrt $\bbvarphi$, the nonconvex problem \eqref{eqn:MSE} is typically  minimized through gradient descent iterations that use backpropagation to efficiently compute the gradient. From a statistical perspective, {this average loss metric} approaches the expected loss if $K$ is large enough. As detailed soon in Remark \ref{rmk:emp_approx}, it does not capture the dispersion of losses or represent the worst-case scenarios such as the maximum loss. As shown by Fig.~\ref{fig:cvar}, the tail of the sample loss distribution is not directly dependent on the average value, and thus minimizing 
{the average loss} does not guarantee the reduction of worst-case prediction losses. Under a small $K$ or high variability of the samples, the worst-case prediction losses can be very large, even if the average is reasonably small. For the voltage optimization problem, these worst-case scenarios may lead to high mismatch in $\bbq^g$ prediction or even severe violations of voltage limits.

To tackle this issue, we propose to develop a risk-aware learning approach by including other statistical measures to improve the safety guarantees of the resultant solutions. To quantify the risk of the sample distribution, one possible measure is the \textit{value-at-risk} (VaR), popularly used in finance for portfolio optimization \cite{rockafellar2000optimization,larsen2002algorithms}. For a given   \textit{significance level} $\alpha \in (0,1)$, the $\alpha$-VaR represents the threshold value for the $(1-\alpha)$-quantile of a random distribution, as indicated by the blue dashed line of Fig.~\ref{fig:cvar}. Hence, reducing the VAR can directly mitigate the worst-case sample losses, but unfortunately is difficult to optimize by using samples due to its lack of smoothness and convexity \cite{rockafellar2000optimization}. Instead, we will consider the \textit{conditional value-at-risk (CVaR)}, a risk measure more widely used as in robust optimization and safe reinforcement learning problems; {see e.g., \cite{gabrel2014recent,chow2015advances,cardoso2019risk}}. 
Intuitively, the $\alpha$-CVaR represents the average over the top $\alpha K$ sample errors, which is thus an upper bound of $\alpha$-VaR, as shown by Fig.~\ref{fig:cvar}. Given $\bbvarphi$, the $\alpha$-CVaR is analytically formed by all $K$ samples, as 
\begin{align} \label{eqn:CVaRn} 
	\gamma_\alpha(\bbvarphi) :=  \frac{1}{\alpha K} \sum_{k=1}^K \ell ( \Phi(\bby_k; \bbvarphi), \bbz_k )  \times  \mathbbm{1}\{\ell ( \Phi(\bby_k; \bbvarphi), \bbz_k )  \geq v \} 
\end{align}
where $v$ is the $\alpha$-VaR while  $\mathbbm{1}(\cdot)$ denotes the indicator function. The CVaR metric can be easily computed after using bisection-typed line search to find $v$. Interestingly, it is equivalent to the following optimization problem, as shown by {\cite{rockafellar2000optimization}} {
\begin{align} \label{eqn:CVaR}
	{\gamma_\alpha(\bbvarphi) := \min_{\beta \in \mathbb R} \left\{\beta + \frac{1}{\alpha K} \sum_{k=1}^K \left[ \ell \left( \Phi(\bby_k; \bbvarphi), \bbz_k \right) - \beta \right]_+ \right\} }
\end{align}
}where the positive projection operator $[a]_+ := \max \{0,a\}$. This is because the optimal $\beta$ to problem \eqref{eqn:CVaR} turns out to be the $\alpha$-VaR. 
{The objective function of \eqref{eqn:CVaR} is convex and piecewise-linear wrt $\beta$. Note that if $\ell \left( \Phi(\bby_k; \bbvarphi), \bbz_k \right)$ is linear wrt $\bbvarphi$, then \eqref{eqn:CVaR} can be recast as a convex linear program by forming the epigraph problem and expressing $\left\{ \ell \left( \Phi(\bby_k; \bbvarphi), \bbz_k \right) - \beta \right\}_{k=1}^K$ as linear inequality constraints. }
The convexity property makes CVaR a popular risk measure, while its gradient estimation is also possible. Here, we remark on the generalizability of using sample-based empirical approximation.

\begin{remark} \emph{(Empirical versus expected)} \label{rmk:emp_approx} Both the 
{average loss and CVaR  in \eqref{eqn:MSE}-\eqref{eqn:CVaRn} are the empirical approximations to their expected-value counterparts, as given by 
\begin{align*}
\check f(\bbvarphi) &=  \mathbb E_{(\bby, \bbz)} \left[ \ell \left( \Phi (\bby; \bbvarphi),\bbz \right) \right], \\
\check \gamma_\alpha (\bbvarphi) &= \mathbb E_{(\bby, \bbz)}  \left[ \ell \left( \Phi (\bby;\bbvarphi),\bbz \right) \Big |  \ell \left( \Phi (\bby;\bbvarphi),\bbz \right) \geq \check v \right]
\end{align*}
}where $\check{v}$ denotes the $\alpha$-{VaR} of the underlying distribution. 
Without knowing the actual distribution, the expected terms are approximated using the sample-based metrics. It is well known such approximation asympotically approaches the expected value as $K\rightarrow \infty$, and similarly for their respective minimizers. For finite $K$, it is also possible to bound the generalization error, or the difference between the minimizers to empirical and expected losses using the so-termed Rademacher complexity defined over the given samples \cite{bartlett2002rademacher}. More recently, the generalization bounds for the empirical CVaR minimizer have been similarly analyzed in \cite{lee2020learning}, showing a fixed (slightly higher) scaling of the Rademacher complexity. Thus, optimizing the empirical CVaR can achieve guaranteed performance in obtaining the desiderata minimizer to  $\check \gamma_\alpha (\cdot)$.
\end{remark}

In addition to predicting $\bbz$, we can also represent the risk associated voltage violation wrt the constraint \eqref{eqn:safe_constraint}. Similarly, we can define the voltage-related CVaR loss as
\begin{align} \label{eqn:CVaRv} 
\gamma_\alpha^v(\bbvarphi) :=  \frac{1}{\alpha K} \sum_{k=1}^K |v_n(\Phi(\bby_k; \bbvarphi))|  \times  \mathbbm{1}\{|v_n(\Phi(\bby_k; \bbvarphi))|  \geq v \} 
\end{align}
where the $\alpha$-VAR threshold $v$ is constructed by the voltage samples corresponding to the predicted reactive power decision $\Phi(\bby_k; \bbvarphi)$. Hence, this loss captures the highest voltage deviations based on the $\bbq^g$ prediction. As voltage depends on the full system input [cf. \eqref{eqn:volt_dev}], minimizing the voltage risk would require the joint training of nodal prediction models $\{\Phi_n(\cdot)\}$. 

We consider a general risk-regularized formulation for learning $\Phi(\cdot; \bbvarphi)$, as {
\begin{align} \label{eqn:risk_regularized}
	\min_{\bbvarphi} &~~ f(\bbvarphi) + \lambda \gamma_\alpha(\bbvarphi)
\end{align}
%
where $\lambda > 0$ is a regularization hyperparameter.} The CVaR term can correspond to either the risk of predicting $\bbz$ in \eqref{eqn:CVaRn} or the voltage risk in \eqref{eqn:CVaRv}, or the combination of the two. This risk-aware learning approach also includes the CVaR-only minimization as a special case with $\lambda \rightarrow \infty$. Nonetheless, the hyperparameter $\lambda$ nicely balances between the risk reduction and learning efficiency. Note that the CVaR function in \eqref{eqn:CVaR} is estimated over  $\alpha K$ samples, which is much smaller than $K$. Accordingly, it can incur high variance of estimation error under a small $\alpha$ value. Hence, purely minimizing CVaR may significantly affect the learning efficiency and potentially lead to much higher expected loss due to the trade-off between mean and variance.   

\subsection{Accelerated Learning via Mini-Batch Selection}
\label{sec:batch_selection}
To solve the risk-aware learning problem \eqref{eqn:risk_regularized}, we adopt the mini-batch gradient descent method \cite{nemirovski2009robust,ghadimi2016mini}, which is widely used in practice for reducing computation complexity and resources. Per iteration $i$, one can randomly select a subset of samples in $\ccalB^i \subset \{1,\ldots, K\}$, and update the NN parameters according to this mini-batch as
\begin{align} \label{eqn:grad_update}
\bbvarphi^{i+1} &= \bbvarphi^i - \eta_\varphi \left( \nabla f(\bbvarphi^i) + \lambda \nabla \gamma_\alpha(\bbvarphi^i) \right)
\end{align}
where $\eta_\varphi$ is a positive step-size (or, learning rate) that is chosen to be sufficiently small for convergence. The step-size can also be adaptively learned for better convergence rates, {using e.g., the ADAM method \cite{goldstein2015adaptive}}. 
Computing the gradient of the average loss in \eqref{eqn:MSE} is the backbone of regular NN training. Similar to the average loss, its gradient can be formed by averaging over the mini-batch $\ccalB^i$, as
\begin{align} \label{eqn:MSEgrad}
\nabla f(\bbvarphi^i) = \frac{2}{|\ccalB^i|} \sum_{k\in \ccalB^i} (\Phi(\bby_k;\bbvarphi^i) - {\bbz}_k)^\top \nabla \Phi(\bby_k; \bbvarphi^i),
\end{align}
by evaluating the sample gradient $\nabla \Phi(\bby_k; \bbvarphi^i)$ using the backpropagation algorithm.

Unfortunately, the gradient for CVaR requires the knowledge of the actual distribution of the sample loss $\ell \left( \Phi(\bby_k; \bbvarphi), \bbz_k \right)$ based on $\bbvarphi^i$ \cite{tamar2015optimizing}, with the worst samples depending on $\bbvarphi$ in a non-smooth fashion.  Hence, a naive approach by truncating the gradient estimates for the worst samples may not be numerically stable. 
One possible solution is to use the equivalent CVaR definition in \eqref{eqn:CVaR}, which does not explicitly use the worst samples. This way, $\beta$ should be included as an optimization variable and updated per iteration $i$, as well. In addition, by approximating $[a]_+$ using the smooth softplus function $ \log(1+e^a)$ as in \cite{tarnopolskaya2010cvar}, one can use the gradient $\nabla_a \log(1+e^a)= 1/(1+e^{-a})$ to evaluate 
\begin{align}
\label{eqn:cvargrad}
\nabla \gamma_\alpha (\bbvarphi^i) = \frac{2}{\alpha |\ccalB^i|} \sum_{k\in\ccalB^i} (\Phi(\bby_k; \bbvarphi^i) - {\bbz}_k)^\top \nabla \Phi(\bby_k; \bbvarphi^i) \nonumber \\
 \frac{1}{1+ \exp (-\|\Phi(\bby_k ;\bbvarphi^i) - {\bbz}_k\|_2^2 + \beta^i)},
\end{align}
In addition, the gradient for $\beta^i$ can be formed to update the auxiliary $\beta$ variable. Similar to \eqref{eqn:MSEgrad}, this smooth CVaR gradient update can be implemented using the backpropagation algorithm. In the numerical tests later on, we will directly use the Pytorch library in Python that directly implements backpropagation along with automatic differentiation (AD) to compute the gradient for CVaR loss. 


Notably, the use of mini-batch $\ccalB^i$ may make the learning efficiency issue more evident for the computation of CVaR's gradient. This is because by randomly selecting a subset of samples, the worst-case samples may not be evenly represented by every mini-batch. This is a known issue for risk-aware learning. Recently, \cite{curi2019adaptive} has proposed an adaptive sampling approach that selects data points more likely to be the worst cases for computing the gradient for CVaR.  

Inspired by this idea, we propose a mini-batch selection scheme for accelerating the learning process under CVaR loss. Intuitively, for a mini-batch $\ccalB^i$ with very small CVaR, it implies that the selected samples in $\ccalB^i$ do not well represent the worst-case scenarios of the full dataset. Therefore, this mini-batch could be disregarded  in the gradient descent update for $\bbvarphi$. This mini-batch selection is  simple to implement, and yet our numerical tests have shown that it can effectively reduce the number of gradient updates and training time. The algorithmic steps are tabulated in Algorithm \ref{alg:min_batch}. {In practice, Algorithm \ref{alg:min_batch} is performed sequentially in each epoch while multiple mini-batches can be generated simultaneously. } Very recently, CVaR based risk-aware learning has been shown to potentially attain linear convergence rates even for non-convex loss functions \cite{kalogerias2020noisy}. We plan to pursue the design of NN architecture that could improve the convergence rate analysis in future.  
\begin{figure}[t]
\vspace*{-5mm}
\begin{algorithm}[H]
	\caption{Mini-batch Selection for Learning \eqref{eqn:risk_regularized}}
	\begin{algorithmic}[1]
		\State \textbf{Input:} Full data $\{(\bby_k, \bbz_k)\}_{k=1}^K$, the NN architecture $\Phi(\cdot;\bbvarphi)$, the hyperparameters $\alpha$, $\lambda$, and $\eta_\varphi$, as well as an iteration stopping threshold $\epsilon$.
		\State \textbf{Output:} The NN parameters in $\bbvarphi$. 
		\State \textbf{Initialize}: Set the iteration index $i=0$ with the initial $\bbvarphi^0$ and CVaR threshold $\gamma_\alpha$. 
		\While{$\|\bbvarphi^i -\bbvarphi^{i-1}\|_2 \geq \epsilon $}
	
        	\State Generate the mini-batch $\ccalB^i$ and compute its CVaR.
			\If{CVaR($\ccalB^i$)$\geq \gamma_\alpha$}
			\State Update the CVaR threshold $\gamma_\alpha\leftarrow$ CVaR($\ccalB^i$).
				\State  Calculate the gradients \eqref{eqn:MSEgrad}-\eqref{eqn:cvargrad} over $\ccalB^i$ using backpropagation.
				\State Update $\bbvarphi^{i+1}$ as in \eqref{eqn:grad_update}.
				\State Update $i\leftarrow i+1$.
			\EndIf
		\EndWhile
		\State \textbf{Return:} the NN model in $\Phi(\cdot;\bbvarphi^i)$
	\end{algorithmic}
	\label{alg:min_batch}
\end{algorithm}
\vspace*{-10mm}
\end{figure}

%% file: Results.tex
\section{Numerical Validations} \label{sec:results}

We have tested on the single-phase equivalent of the IEEE 123-bus test case \cite{ieee123} to demonstrate the effectiveness of the proposed risk-aware learning algorithm in mitigating the risks of $\bbq^g$ prediction and voltage violation. The test system consists of 90 load nodes with nodes 66, 85, 96, 114, 151, and 250 equipped with inverter-based PV generation. Real-world active power data for PV output $\bbp^g$ and non-PV loads $\bbp^c$ at the minute-level resolution has been obtained from the Pecan Street Dataport \cite{pecan}. The reactive power $\bbq^c$ for the loads was synthetically generated by randomly selecting a power factor in the range $[0.9,~0.95]$. The optimal reactive power $\bbz$ in \eqref{eqn:LCQP} has been solved  for each sample system OC $\bby$ using the MATLAB\textsuperscript{\textregistered} R2020b. 
10 days of data have been used to generate 14,400 samples. 
{Samples from the first 8 days are used to train the NN models in a batch setting, from which each mini-batch is randomly generated. The remaining 2 days are used to test the trained models in predicting the reactive power decisions from nodal measurements.}
	

To attain scalable decision rules, all six PV nodes use the same NN model with the input $\bby_n$ consisting of p/q data locally and from the feeder head's broadcast. To allow for efficient training, a simple graph NN model with the graph filter being an identity matrix (e.g., no information sharing among nodes) \cite{liu2021graph} and \texttt{relu} activation function has been used to set up $\Phi(\cdot)$ for this local architecture. 

{To evaluate the performance of the proposed algorithm, we have considered three loss objectives: (i) the (risk-neutral) MSE-only loss \eqref{eqn:MSE} for predicting $\bbq^g$; 
{(ii) the risk-regularized MSE \eqref{eqn:risk_regularized} with CVaR on the error in predicting  $\bbq^g$, denoted as CVaR(qg);}
{(iii) and the risk-regularized MSE with CVaR on both  $\bbq^g$ prediction error and voltage deviation, denoted as CVaR(qg,dv).} The CVaR parameter $\alpha=0.2$ has been picked.  For both CVaR-based risk-aware objectives, we have compared Algorithm \ref{alg:min_batch} with the default mini-batch based algorithm. These algorithms have been implemented using the PyTorch library and tested on Google Colaboratory using the NVIDIA Tesla P100 GPU for training acceleration. {All models converged in the training process, and the same convergence criterion for the loss is applied for all test cases.} All results presented here have used the test set to generate the error performance\footnote{The codes and results are available at: \newline \url{https://github.com/ShaohuiLiu/RiskAwareLearning_VoltageOpt_DistGrid}}.}

\begin{figure}[tb!]
\vspace*{-5mm}
\centering
\subfigure{\label{fig:q_noq}\includegraphics[width=87mm]{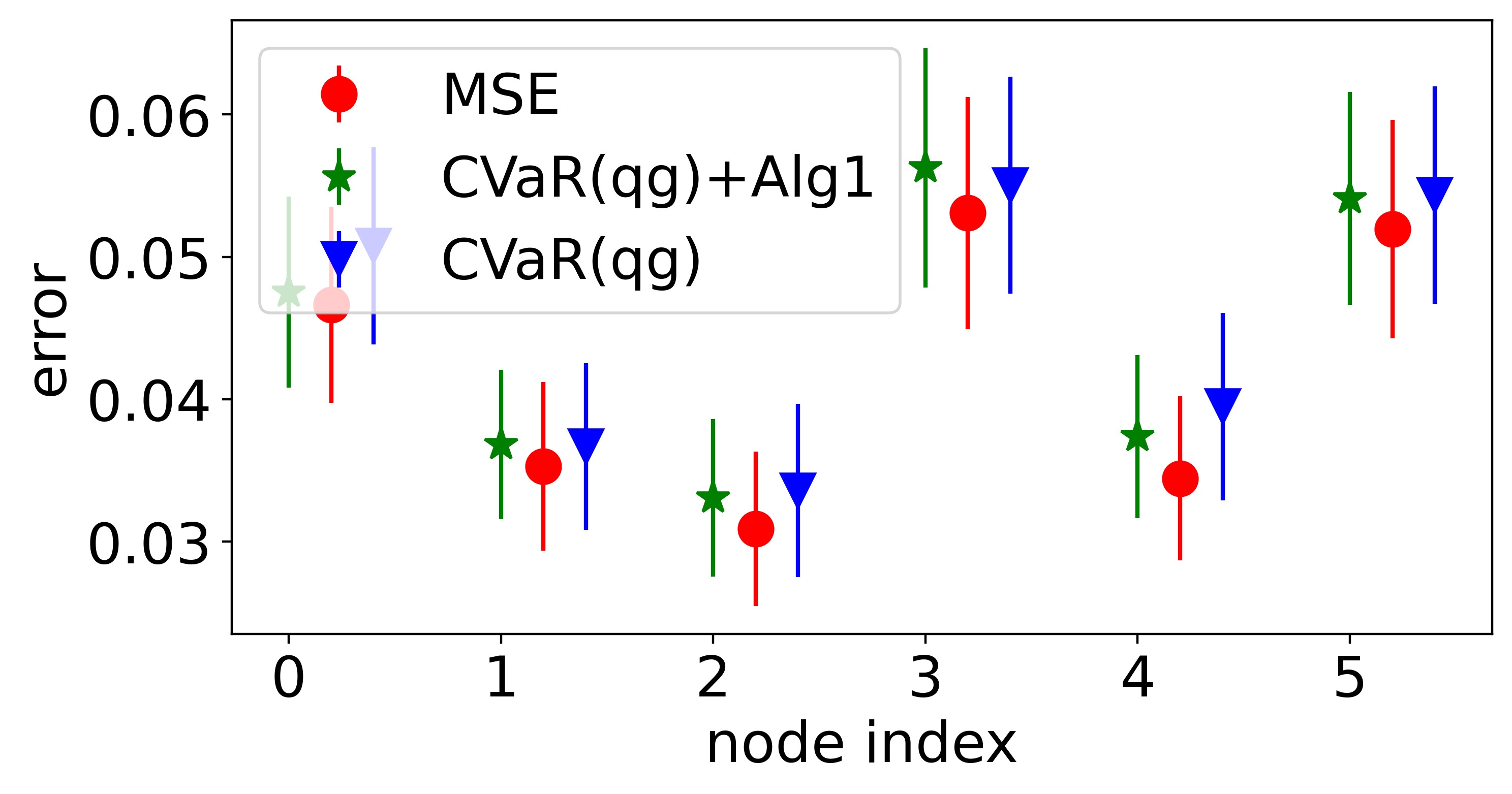}
\vspace*{-30mm}}
\subfigure{\label{fig:dv_noq}\includegraphics[width=87mm]{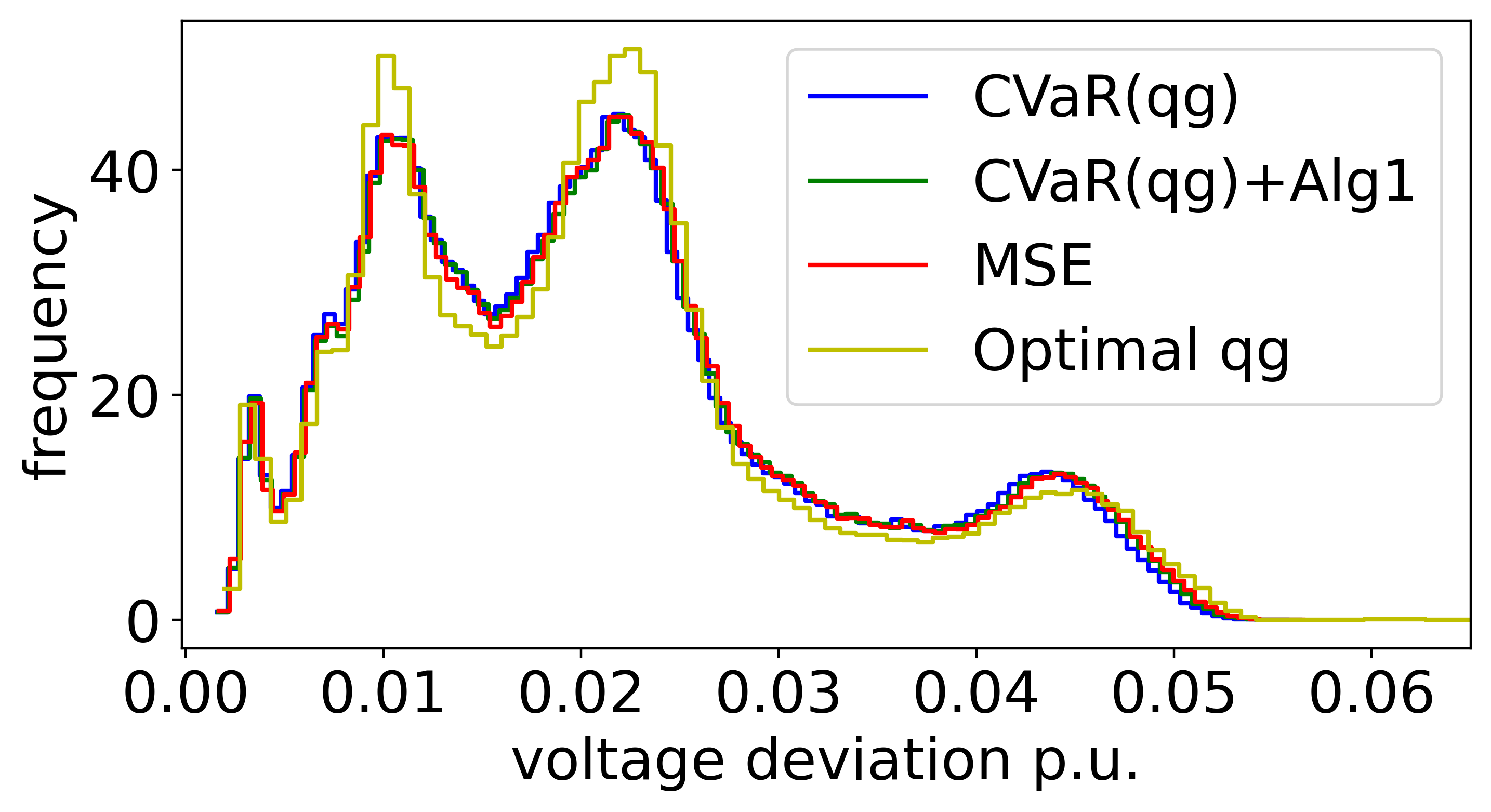}} 
\vspace*{-3mm}
\caption{{{\bf Test Case 1:} Comparisons of the three NN training approaches in terms of (top) nodal error in predicting $q^g_n$ (mean $\pm$ standard deviation) and (bottom) the distribution of voltage deviations based on the training results.}}
\label{fig:training_err_case1}
\end{figure}

\begin{table}[tb!]
\vspace*{+2mm}
\centering
\caption{{Computation time and error performance in Test Case 1.}}
\begin{tabular}{lrrrr}
\hline
Loss obj. & \multicolumn{1}{l}{Epoch [s]} & \multicolumn{1}{l}{Total [s]} & \multicolumn{1}{l}{$q^g$ error} & \multicolumn{1}{l}{Max $\bbv$} \\ \hline
MSE      & 0.52                      & 46.48                     & 6.62\%      &5.65\%                          \\
CVaR(qg)     & 1.07                      & 38.70                     & 6.78\%                           & 5.57\%                          \\
CVaR(qg)+Alg 1 & 0.61                       & 35.63                     & 6.97\%                           & 5.64\%                          \\ \hline
\end{tabular}
\label{table:case1}
\vspace*{-5mm}
\end{table}

{{\textbf{Test Case 1:}} We first compare the training objectives (i) MSE and (ii) CVaR(qg) to demonstrate the benefits of incorporating the CVaR loss for predicting $\bbq^g$, while the CVaR(qg) has also been implemented by the proposed Algorithm 1. Fig.~\ref{fig:training_err_case1} and Table \ref{table:case1} list the training time and test error performances. Fig.~\ref{fig:training_err_case1} plots the nodal prediction error (top)  and the corresponding distribution of voltage deviations (bottom). In general, the error performance {in both $\bbq^g$ prediction and voltage deviation} is very close among the three methods, with the voltage distribution very similar to that of the optimal decisions. The CVaR loss has slightly increased the prediction error, also confirmed by the percentage prediction error and maximum voltage deviation in Table \ref{table:case1}. This change is because the task of predicting $\bbq^g$ has been very accurate using the input features to the NN training. Interestingly, although the CVaR loss increases the average computation time per epoch, Table \ref{table:case1} shows that this regularization actually speeds up the overall training process as comparing to the MSE loss. More importantly, the proposed Algorithm \ref{alg:min_batch} has attained the expected CVaR error performance while reducing the computation time for each epoch (by over $40\%$) and the total training time. Thus, the proposed Algorithm \ref{alg:min_batch} has attained faster learning speed for predicting $\bbq^g$. }

\begin{figure}[tb!]
\vspace*{-5mm}
\centering
\subfigure{\label{fig:q_wq}\includegraphics[width=87mm]{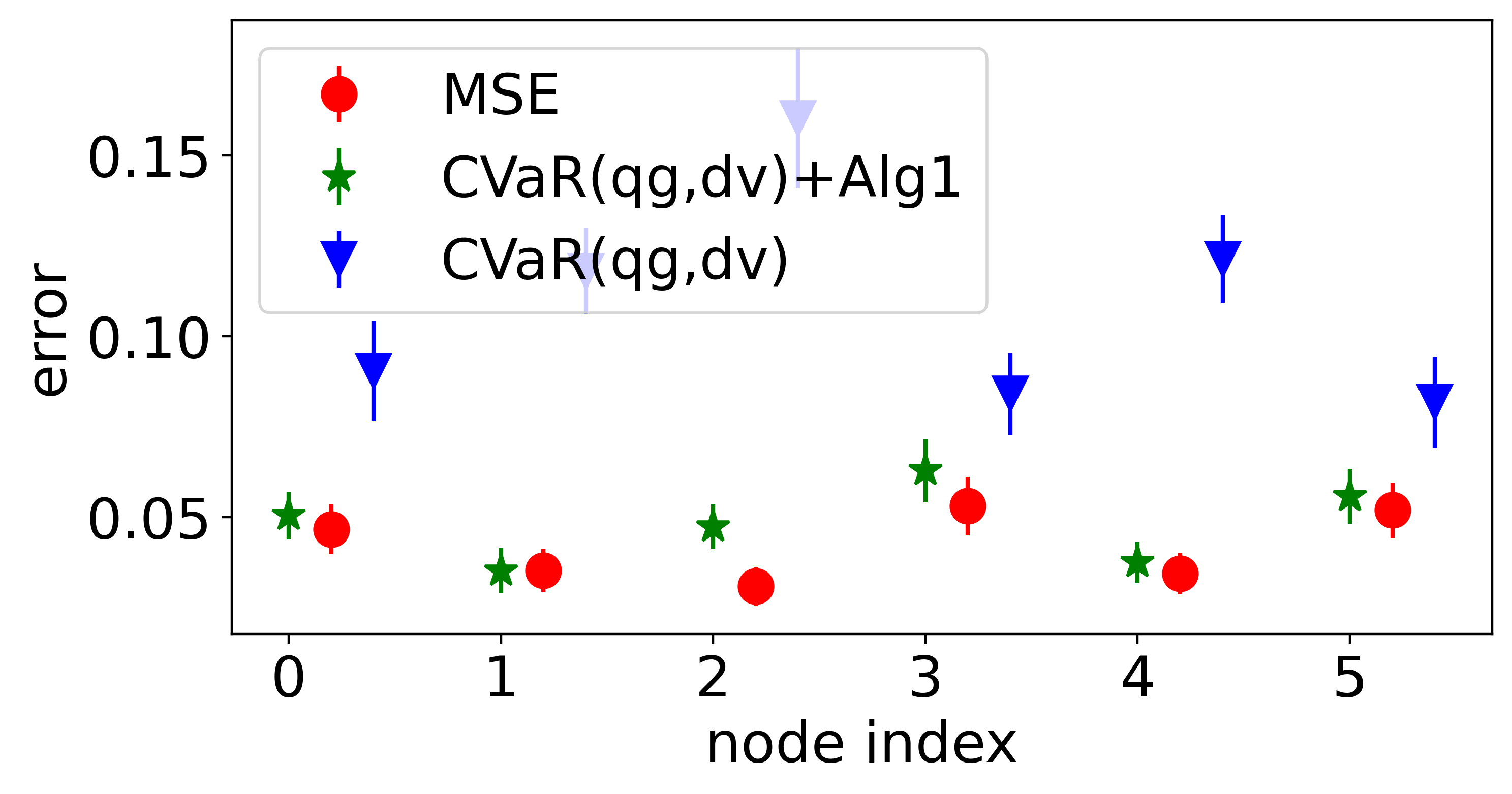}
\vspace*{-20mm}}
\subfigure{\label{fig:dv_wq}\includegraphics[width=87mm]{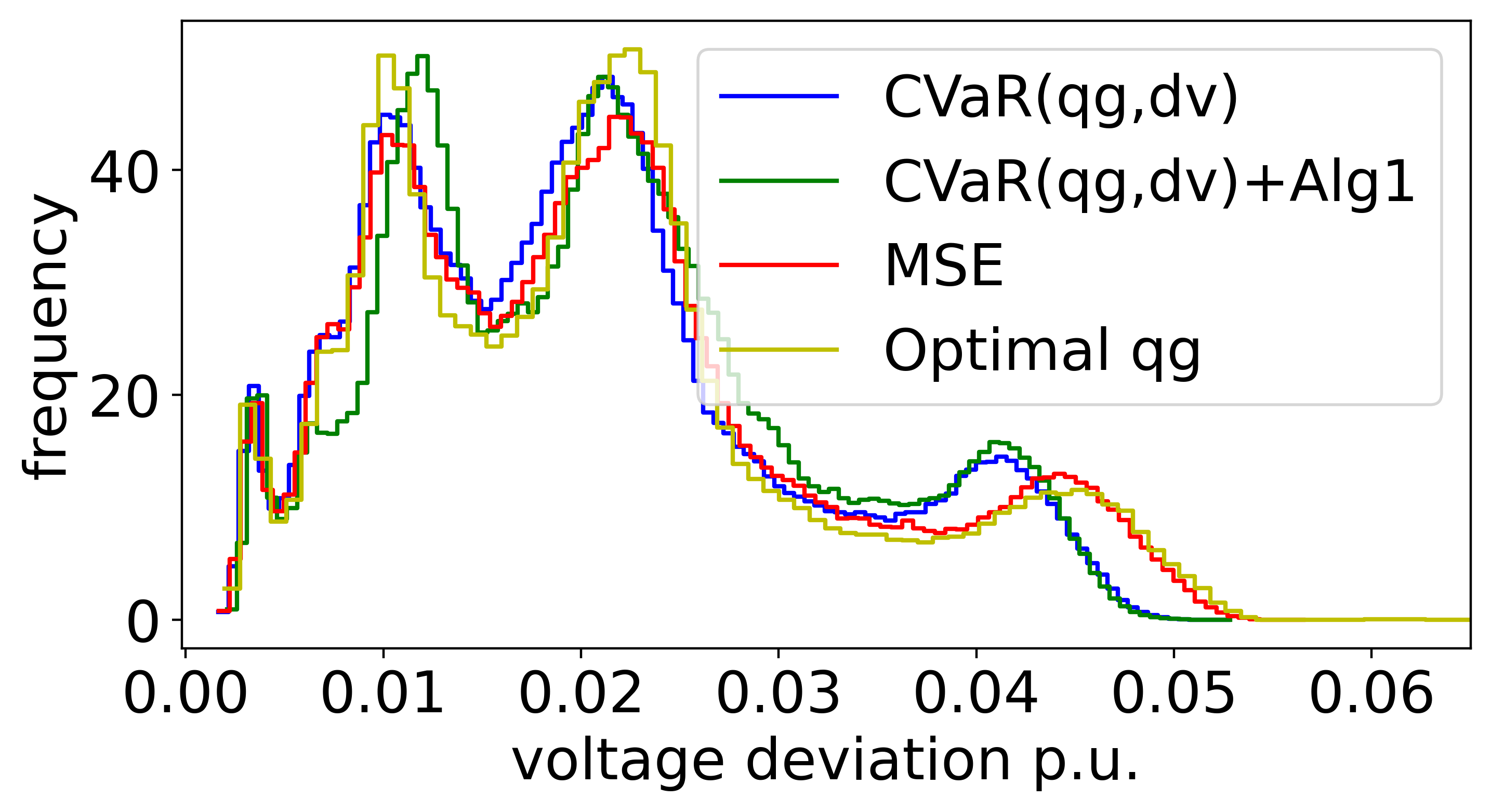}} 
\vspace*{-3mm}
\caption{{{\textbf{Test case 2} with CVaR regularization on both $\bbq^g$ and voltage deviation $\bbv$:} Comparison of the NN-based decentralized controller trained using pure MSE loss, the CVaR-regularized formulation as well as the CVaR-regularized controller trained using CVaR-based mini-batch selection algorithm on $\bbq^g$ in terms of (top figure) predicting $\bbq^g$ (mean $\pm$ standard deviation) and (bottom) the resulting histogram of voltage deviations.}}
\vspace*{-5mm}
\label{fig:training_err_case2}
\end{figure}

\begin{table}[tb!]
\vspace*{-5mm}
\centering
\caption{{Computation time and error performance in Test Case 2.}}
\begin{tabular}{lrrrr}
\hline
Loss obj. & \multicolumn{1}{l}{Epoch [s]} & \multicolumn{1}{l}{Total [s]} & \multicolumn{1}{l}{$q^g$ error} & \multicolumn{1}{l}{Max $\bbv$} \\ \hline
MSE      & 0.54          & 44.89     & 4.21\% & 5.65\% \\
CVaR(qg,dv)     & 0.77    & 31.73      & 10.93\%      & 5.27\%        \\
CVaR(qg,dv)+Alg 1 & 0.51       & 25.93      & 4.82\%        & 5.28\%     \\ \hline
\end{tabular}
\vspace*{-5mm}
\label{table:case2}
\end{table}

{\textbf{Test Case 2:} We further incorporate the CVaR loss associated with voltage deviation by comparing the training objectives (i) MSE and (iii) CVaR(qg,dv). Fig.~ \ref{fig:training_err_case2} and Table \ref{table:case2}  list the updated training time and test error performance for this comparison. Notably, Fig.~\ref{fig:training_err_case2} demonstrates that the CVaR loss introduces additional difficulty to optimize, leading to higher nodal prediction error over the MSE one. Thanks to the proposed acceleration scheme, Algorithm \ref{alg:min_batch} has significantly mitigated the error bias issue by using more statistically important mini-batches. More importantly, using the additional voltage-risk \eqref{eqn:CVaRv}, the proposed method can reduce the maximum voltage deviation by around $7\%$  over the MSE one,  as shown in Table \ref{table:case2}.  This is more evident in Fig.~\ref{fig:training_err_case2} where the worst-case voltage deviations (over 0.05) are effectively reduced in frequency {thanks to the CVaR regularization on the worst-case voltage. Thus, the voltage-based CVaR metric is especially useful for improving the safety of the resultant decision rules}. In addition, the training time improvement is more significant in this test case, with the total training time reduced by over $40\%$ from the MSE one.}
Hence, the proposed 
risk-constrained learning framework for designing decentralized controllers has shown to be effective in attaining safe decision making for DERs to perform voltage optimization. To sum up, the proposed CVaR regularization and mini-batch selection scheme can effectively improve the training speed, while incorporating the voltage risk can help mitigate the worst-case voltage deviations attained by the NN-based decision rules.

%% file: Conclusion.tex
\section{Conclusions and Future Work} \label{sec:conclusion}

This paper developed a risk-aware learning framework for attaining scalable decision rules in the distribution grid voltage optimization problem. For learning the optimal reactive power decision rules using local data, we propose to account for the worst-case scenarios by considering the CVaR losses associated with prediction error and voltage deviation. To solve the resultant risk-regularized problem, we develop a mini-batch gradient descent algorithm by judiciously selecting the mini-batches to accelerate the training process. Numerical tests using real-world data have demonstrated the training accelerations by using the proposed mini-batch selection algorithm. In addition, the benefits of using voltage-associated risk have been validated in terms of mitigating the worst-case voltage deviations.

Several interesting future directions open up for this work. We are currently investigating the convergence properties of optimizing CVaR loss. In addition, it is interesting to incorporate the graph structure of distribution grids to generalize the scalable NN architecture {and to investigate the effects of system topology on our proposed approach}. Last, the proposed risk-aware learning framework can be extended to optimize active power resources and also dynamical DERs such as energy storage.